\documentclass{article}
\usepackage{spconf,amsmath,graphicx,hyperref}
\usepackage{amsmath}  
\usepackage{amssymb}  
\usepackage{graphicx}  
\usepackage{algorithm}
\usepackage{algpseudocode}
\usepackage{amsmath}
\usepackage{booktabs}
\usepackage{xcolor}
\usepackage{enumitem}
\usepackage{graphicx} 
\usepackage{amsmath}  
\usepackage{multirow} 

\title{Balancing Rewards in Text Summarization: Multi-Objective Reinforcement Learning via HyperVolume Optimization}
\name{Junjie Song\sthanks{Equal Contribution}, 
Yiwen Liu$^*$,
Dapeng Li$^*$,
Yin Sun,
Shukun Fu,
Siqi Chen,
Yuji Cao\sthanks{Corresponding Author}
}
\address{AI-4-Business of Li Auto Inc.  \\ Beijing, China}

\begin{document}
%
\maketitle
\begin{abstract}
Text summarization is a crucial task that requires the simultaneous optimization of multiple objectives, including consistency, coherence, relevance, and fluency, which presents considerable challenges. Although large language models (LLMs) have demonstrated remarkable performance, enhanced by reinforcement learning (RL), few studies have focused on optimizing the multi-objective problem of summarization through RL based on LLMs. In this paper, we introduce hypervolume optimization (HVO), a novel optimization strategy that dynamically adjusts the scores between groups during the reward process in RL by using the hypervolume method. This method guides the model’s optimization to progressively approximate the pareto front, thereby generating balanced summaries across multiple objectives.
Experimental results on several representative summarization datasets demonstrate that our method outperforms group relative policy optimization (GRPO) in overall scores and shows more balanced performance across different dimensions. Moreover, a 7B foundation model enhanced by HVO performs comparably to GPT-4 in the summarization task, while maintaining a shorter generation length. Our code is publicly available at \url{https://github.com/ai4business-LiAuto/HVO.git}
\end{abstract}
\section{Introduction}
\label{sec:intro}
Text summarization is a core and challenging task in natural language processing (NLP)~\cite{DBLP:journals/csur/ZhangYZ25}. To comprehensively evaluate the quality of generated summaries, researchers typically examine multiple dimensions, such as coherence, consistency, fluency, and relevance~\cite{DBLP:conf/emnlp/LiuIXWXZ23}. 
However, optimizing the objectives of these dimensions simultaneously is challenging, as improvements along one dimension may lead to compromises in others~\cite{abo2023automatic}, resulting in imbalanced summaries.
\begin{figure}[htb]
\begin{minipage}[b]{1\linewidth}
  \centering
  \centerline{\includegraphics[width=1\textwidth]{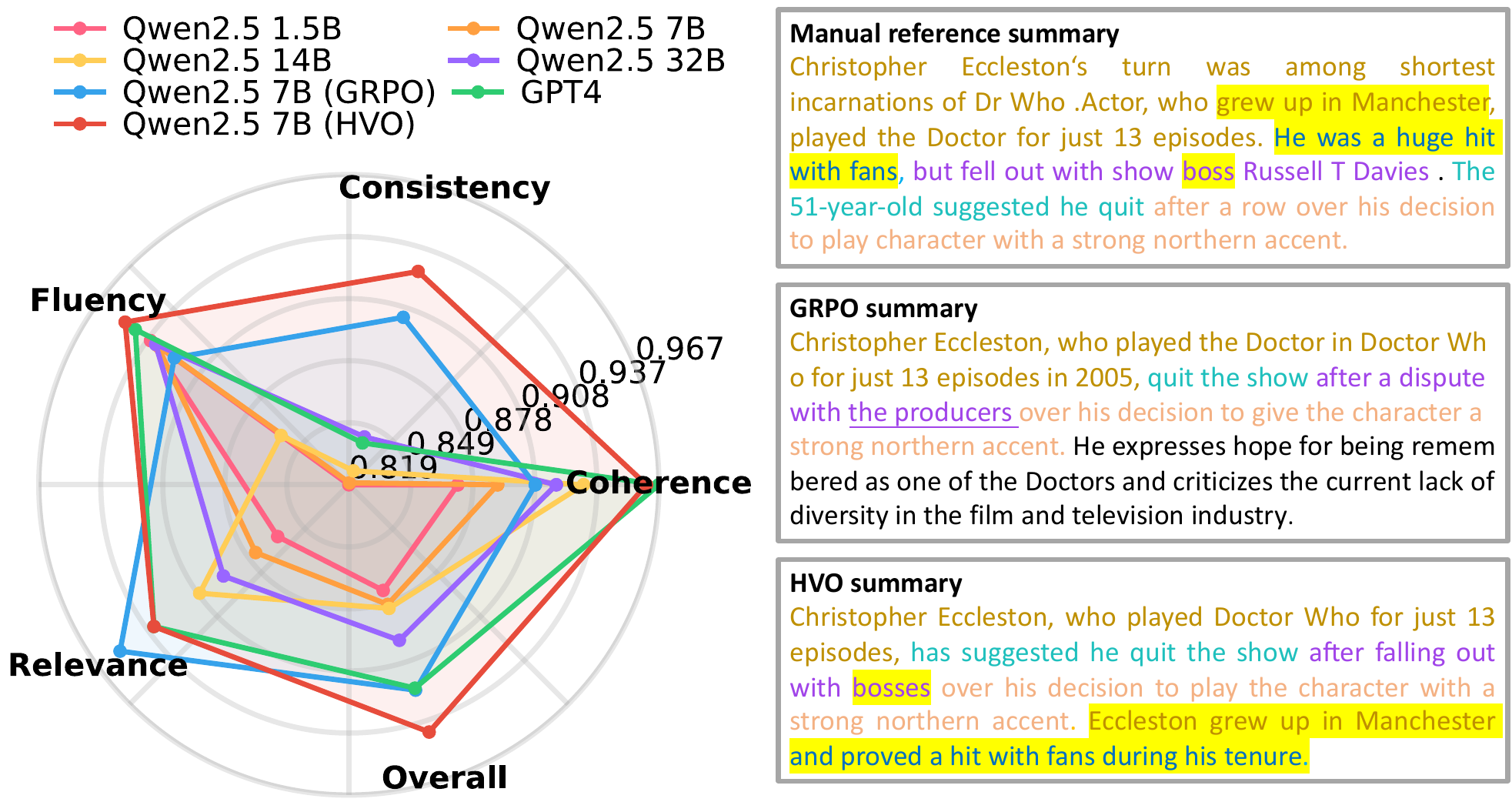}}
\end{minipage}
\caption{The radar chart shows the scores of different models in each dimension, evaluated by UniEval~\cite{DBLP:conf/emnlp/Zhong0YMJLZJH22} on the CNN/DailyMail~\cite{nallapati-etal-2016-abstractive}. The same meanings are represented in the same color and the highlighted areas show where the HVO summary performs better. Underline indicates improper phrasing.}
\label{fig:cnn_rader}
\vspace{-15pt}  
\end{figure}

With the development of large language models (LLMs), utilizing LLMs for zero-shot generation of summaries has become one of the mainstream approaches~\cite{DBLP:journals/corr/abs-2309-09558}. As shown in Figure~\ref{fig:cnn_rader}, the zero-shot summaries leave room for improvement in all dimensions, especially in consistency. With the demonstrated success of reinforcement learning (RL) in post-training~\cite{li2025efficient,li2024verco}, it has been established as an effective enhancement method in text summarization~\cite{DBLP:journals/csur/ZhangYZ25}. Most studies use RL methods to enhance summarization models by relying on a single reward signal~\cite{DBLP:conf/acl/YadavGAD20,DBLP:journals/nle/FikriOY24,roit2023factually}. However, they do not integrate multi-dimensional metrics as rewards for summarization, and those methods still encounter challenges in generating high-quality summaries while balancing multiple dimensions. To achieve well-balanced summaries, MDO~\cite{DBLP:conf/acl/RyuD0LO24} incorporates multi-dimensional rewards and explores optimal strategies for multi-objective RL. In detail, MDO uses PCGrad optimization to reduce gradient interference across different dimensions, facilitating the discovery of Pareto improvements by balancing multiple objectives. However, this method requires pairwise gradient projections between different dimensions. Due to the high computational cost, it is not feasible to integrate into LLMs.


In this work, we propose hypervolume optimization (HVO), a multi-objective reinforcement learning (MORL) optimization strategy designed for text summarization, which is based on group
relative policy optimization (GRPO)~\cite{shao2024deepseekmathpushinglimitsmathematical} and well-suited for LLMs. HVO incorporates multi-dimensional rewards into a hypervolume~\cite{guerreiro2020hypervolume} computation framework, optimizing the model towards hypervolume maximization, progressively
approaching the Pareto optimal frontier. As illustrated in Figure~\ref{fig:cnn_rader}, our proposed HVO method significantly outperforms LLM-based baseline approaches by achieving higher scores across multiple evaluation dimensions, while maintaining a more balanced and stable optimization. The main contributions of this work are summarized as follows:

\begin{itemize}[noitemsep, topsep=0pt]
\item We introduce HVO, a multi-objective reinforcement learning strategy for text summarization based on
GRPO, which efficiently balances multiple evaluation
dimensions without requiring supervised fine-tuning or
a cold start.

\item On representative datasets, HVO outperforms GRPO with better hypervolume and UniEval scores. The 7B LLM version of HVO performs similarly to GPT-4 on two benchmarks.

\item To address the training instability and summary length collapse issues in vanilla GRPO, a new length constraint mechanism is introduced to enhance the training stability.
\end{itemize}

\section{METHOD}
\label{sec:intro}

In order to preserve the core summarization capabilities of LLMs while enhancing performance across multiple dimensions. We introduce HVO based on the R1-Zero-like~\cite{deepseekai2025deepseekr1incentivizingreasoningcapability} training paradigm, which directly applies GRPO to base LLMs without relying on supervised fine-tuning (SFT) as a preliminary step. Specifically, HVO applies hypervolume evaluation to improve the multi-objective optimization process, where a policy model generates summaries optimized with multi-dimensional rewards calculated via UniEval~\cite{DBLP:conf/emnlp/Zhong0YMJLZJH22} , enhanced by a length constraint mechanism. UniEval is a multi-dimensional evaluation tool that strongly correlates with human judgment~\cite{DBLP:conf/emnlp/Zhong0YMJLZJH22} and is commonly used in text summarization research~\cite{ DBLP:conf/acl/RyuD0LO24, DBLP:conf/icassp/JungSCR024}. The entire process is illustrated in Figure \ref{fig:structure}.

\subsection{Problem Formulation}

Given a set of documents $\{p_1, p_2, \dots, p_N\}$, LLMs generate the corresponding summaries $\{s_1, s_2, \dots, s_N\}$ using a prompt $\alpha$. Text summarization can be modeled as a Question Answering (QA) problem, where the question is a prompted document $q_i=\textit{f}(p_i;\alpha) $ and the answer is the corresponding summary $s_i$. The dataset $\mathcal{D}$ is formed as $\{(q_i, s_i)\}_{i=1}^N$. For a specific question $q \in \mathcal{D}$
the policy model $\pi_\theta$ generates a group of $G$ individual
summaries $\{o_i\}_{i=1}^G$. For each evaluation dimension $D_k$ in $\{D_1, D_2, \dots, D_M\}$, the reward model $R$ computes the reward $r^{k}_i$ for $o_i$ at the 
$k$-th dimension. Our objective is to optimize the policy model's parameters $\theta$ using multi-objective reinforcement learning to maximize the expected reward guided by hypervolume maximization based on GRPO. 
\subsection{HVO}
\begin{figure}[ht]
  \centering
  \includegraphics[width=0.5\textwidth]{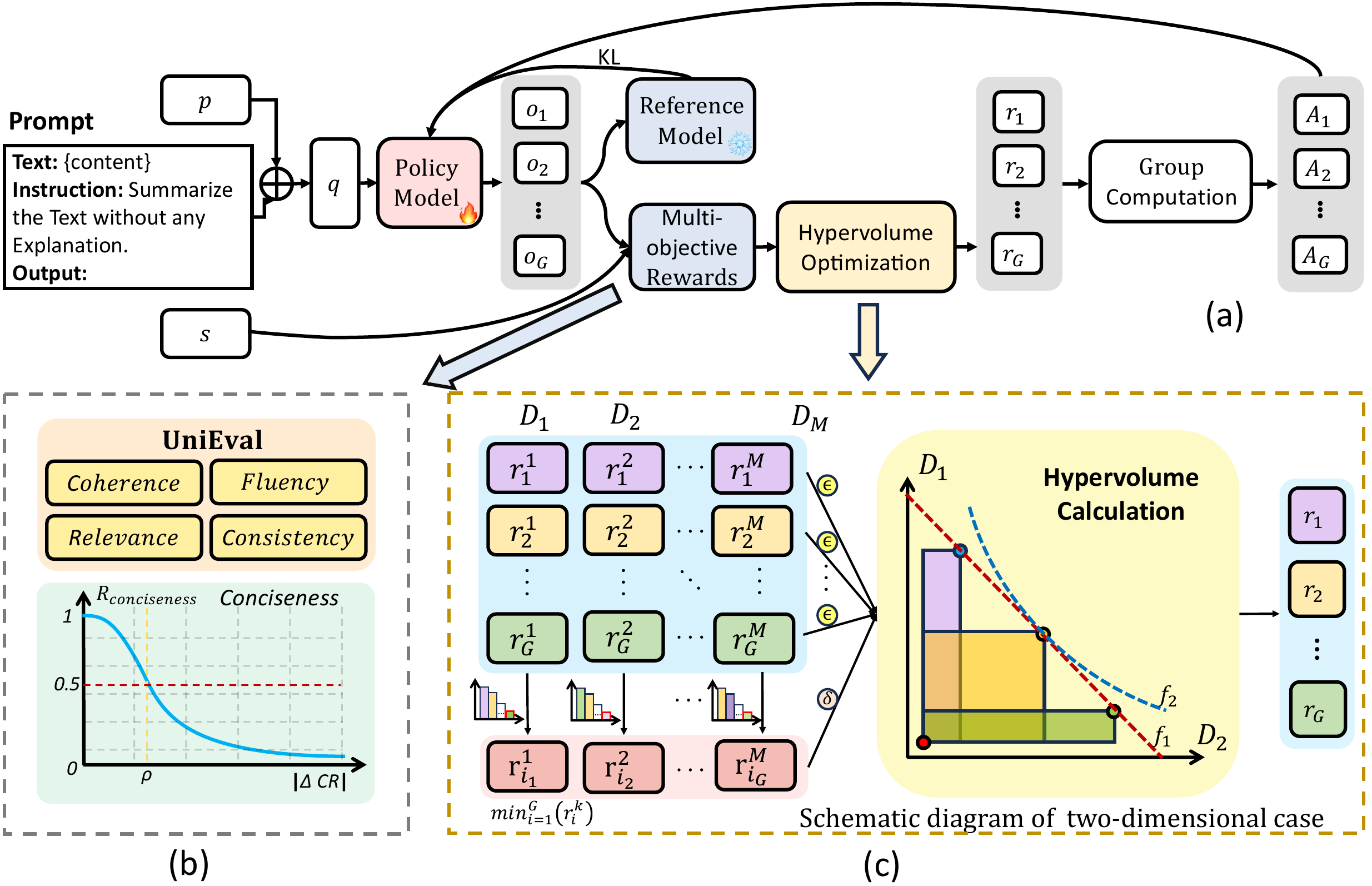}
  \caption{The entire process of HVO. In subplot (c), the points on the $f_1$ line represent the same linear weighted sum score for $D_1$ and $D_2$, while the points on the $f_2$ line represent the same hypervolume value for $D_1$ and $D_2$.}
  \label{fig:structure}
\end{figure}

HVO is based on GRPO in which the optimization of policy $\pi_\theta$ is achieved by maximizing the following objective function:
\begin{align}
\mathcal{J}(\theta) &= \mathbb{E}_{(q, a) \sim \mathcal{D}, \{o_i\}_{i=1}^G \sim \pi_{\text{old}}\nonumber (\cdot | q)} \Bigg[ \frac{1}{G} \sum_{i=1}^G \frac{1}{|o_i|} \sum_{t=1}^{|o_i|} \Bigg( \\
& \min \left( f_{i,t}(\theta) \hat{A}_{i,t}, \text{clip}( f_{i,t}(\theta), 1-\epsilon, 1+\epsilon ) \hat{A}_{i,t} \right) \\
& - \beta D_{\text{KL}}(\pi_{\sigma} || \pi_{\text{ref}}) \Bigg) \Bigg], \nonumber
\end{align}
where
\[
\begin{aligned}
f_{i,t}(\theta) &= \frac{\pi_{\theta}(o_{i,t} | q, o_{i,<t})}{\pi_{\theta_{\text{old}}}(o_{i,t} | q, o_{i,<t})}, \quad
\hat{A}_{i,t} &= \frac{r_i - \text{mean}(\{r_i\}_{i=1}^G)}{\text{std}(\{r_i\}_{i=1}^G)}.
\end{aligned}
\]
The default and direct way to extend GRPO to multi-objective optimization is to compute the reward \( r_i \) through a weighted linear combination:
\begin{align}
r_i &= \sum_{k=1}^{M} w^{k} \cdot r_{i}^{k},
\end{align}
where \( w^k \) represents the weight for the reward $r_{i}^{k}$, which requires manual configuration. It is worth noting that while the weighted linear combination method is simple, it has non-trivial limitations, particularly in handling issues of inter-dependencies between objectives, which can result in imbalanced or incomplete optimization outcomes~\cite{das1997closer}.

The hypervolume method is an evaluation metric in multi-objective optimization that measures the volume of the hypercube occupied by a set of solutions in the objective space. As shown in Figure \ref{fig:structure} (c), taking the two-dimensional case as an example, when the weighted linear combination scores of the samples are similar, samples with more balanced dimensions have higher hypervolume values. Moreover, hypervolume-based evaluation has been shown to be a Pareto-consistent evaluation method~\cite{DBLP:journals/tec/ZitzlerTLFF03}. Therefore, optimizing towards hypervolume maximization ensures continuous improvement in the quality of the solution set, progressively approaching the Pareto optimal frontier. Inspired by this, we integrate hypervolume evaluation into the multi-dimensional rewards of GRPO and use the commonly adopted approach of selecting a slightly worse reference point than the nadir point. The specific method of HVO is as follows:
\begin{align} r_i &= \prod_{k=1}^{M} \left[ \min \left( \epsilon, r_i^k - \min(\{r_i^k\}_{i=1}^{G}) + \delta \right) \right ]^{-w^k},
\label{eq:hypervolume}
\end{align}
where \( \delta, \epsilon \in (0, 1) \),  $\delta$ is a small constant used to avoid zero values and
$\epsilon$ is an upper bound, which ensures that the term \( r^k_{i} - \min_{i=1}^{G}(r^k_{i}) \) is restricted within the range \( [\delta, \epsilon] \), allowing monotonic adjustment of $w^k$ for $D_k$. 
\renewcommand{\arraystretch}{0.80}

\begin{table*}[ht]
\centering
\setlength{\tabcolsep}{4pt}
{\normalsize
\begin{tabular}{c |r  r|  c c c c c c c  }
\toprule
\textbf{Dataset}&\textbf{Model} & \textbf{Method} & \textbf{Coherence} & \textbf{Consistency} & \textbf{Fluency} & \textbf{Relevance} & \textbf{HV score} & \textbf{Overall}  & \textbf{STD}\\
 \hline
  \multirow{8}{*}{\rotatebox{90}{\textbf{BillSum}}} &PEGASUS & SFT & 0.823 & 0.832 & 0.849 & 0.814 & 0.171 & 0.830 & \textcolor{red}{\textbf{0.015}} \\
  &Qwen2.5 1.5B & Zero-shot & 0.941 & 0.826 & 0.905 & 0.914 & 1.018 & 0.896 & \textcolor{blue}{\underline{0.050}} \\
  &Qwen2.5 7B & Zero-shot & 0.958 & \textcolor{blue}{\underline{0.843}} & 0.820 & 0.948 & 0.790 & 0.892 & 0.071 \\
  &Qwen2.5 14B & Zero-shot & 0.953 & 0.805 & 0.919 & 0.940 & 1.101 & 0.904 & 0.068 \\
  &Qwen2.5 32B & Zero-shot & 0.948 & 0.799 & \textcolor{red}{\textbf{0.941}} & 0.933 & 1.088 & 0.905 & 0.071 \\
  &GPT4 & Zero-shot & \textcolor{red}{\textbf{0.973}} & \textcolor{blue}{\underline{0.843}} & 0.831 & \textcolor{red}{\textbf{0.971}} & 1.025 & 0.904 & 0.078 \\
  &Qwen2.5 7B & GRPO & 0.959 & 0.823 & 0.912 & \textcolor{blue}{\underline{0.955}} & \textcolor{blue}{\underline{1.358}} & \textcolor{blue}{\underline{0.912}} & 0.063 \\
  &Qwen2.5 7B & HVO & \textcolor{blue}{\underline{0.964}} & \textcolor{red}{\textbf{0.853}} & \textcolor{blue}{\underline{0.939}} & \textcolor{blue}{\underline{0.955}} & \textcolor{red}{\textbf{1.961}} & \textcolor{red}{\textbf{0.928}} & {{0.051}} \\
\midrule
 \multirow{8}{*}{\rotatebox{90}{\textbf{CNN/DailyMail}}}&PEGASUS & SFT & 0.936 & \textcolor{red}{\textbf{0.939}} & 0.815 & 0.684 & 0.364 & 0.843 & 0.121 \\
 &Qwen2.5 1.5B & Zero-shot & 0.871 & 0.819 & 0.936 & 0.861 & 0.612 & 0.872 & 0.048 \\
 &Qwen2.5 7B & Zero-shot & 0.890 & 0.820 & 0.932 & 0.874 & 0.757 & 0.879 & 0.046 \\
 &Qwen2.5 14B & Zero-shot & 0.931 & 0.826 & 0.859 & 0.907 & 0.805 & 0.881 & 0.047 \\
 &Qwen2.5 32B & Zero-shot & 0.918 & 0.843 & 0.933 & 0.893 & 1.226 & 0.897 & 0.040 \\
 &GPT4 & Zero-shot & \textcolor{red}{\textbf{0.967}} & 0.840 & \textcolor{blue}{\underline{0.945}} & \textcolor{blue}{\underline{0.934}} & 1.913 & 0.921 & 0.056 \\
 &Qwen2.5 7B & GRPO & 0.908 & 0.903 & 0.922 & \textcolor{red}{\textbf{0.954}} & \textcolor{blue}{\underline{1.938}} & \textcolor{blue}{\underline{0.922}} & \textcolor{blue}{\underline{0.023}} \\
 &Qwen2.5 7B & HVO & \textcolor{blue}{\underline{0.961}} & \textcolor{blue}{\underline{0.926}} & \textcolor{red}{\textbf{0.951}} & \textcolor{blue}{\underline{0.934}} & \textcolor{red}{\textbf{3.258}} & \textcolor{red}{\textbf{0.943}} & \textcolor{red}{\textbf{0.016}} \\ \bottomrule 

\end{tabular}
}
\caption{The results of multi-dimensional evaluation measured on both the CNN/DailyMail and BillSum datasets. Within the same dimension, the bold denotes the highest score, and the underline denotes the second-highest score. The HV score is expressed in units of $10^{-3}$ calculated according to Equation \ref{eq:hypervolume}.
}
\label{tab:combined_results}
\end{table*}

We use the scores of UniEval as a reward that focuses on coherence, consistency, fluency, and relevance in text summarization. However, recent studies have shown that GRPO encounters issues with training instability~\cite{yu2025dapo}, which leads to issues with summary length collapse during training when using UniEval as the sole reward. Consequently, we propose a new length constraint method to help maintain training stability. The calculation is as follows:
\begin{align}
R_{\text{conciseness}}\left( o_i\right) &= \frac{1}{1 + \left( x_i/\rho \right)^{\lambda}},
\end{align}
where $
x_i = \left| \left|p_i\right|/\left|o_i\right| - \text{mean}\left( \left\{ \left|p_j\right|/\left|s_j\right| \right\} _{j=1}^{V} \right) \right|
$, $V$ is the size of the training set.
It represents the absolute difference between the compression ratio (CR) of the generated summary $o_i$ and the average compression ratio of human-generated summaries in the training set.
The $\lambda$ represents the steepness, indicating the degree of rapid decrease, while $\rho$ represents the offset, indicating how far from $x_i = 0$ the value starts to decrease sharply. This ensures that the score decreases slowly around $x_i = 0$, maintaining sufficient exploration space, and declines rapidly when $x_i$ is large.

\section{Experiment}
\subsection{Experiment Setup}
\subsubsection{Dataset and Baselines}
We use two text summarization datasets: CNN/DailyMail~\cite{nallapati-etal-2016-abstractive} for news summarization, with 287K training and 11.5K test samples, and BillSum~\cite{kornilova-eidelman-2019-billsum} for legislative content, with 18.9K training and 3.2K test samples.

We used the instruct version of the Qwen 2.5 as the baseline, as it is known for its strong performance in various benchmarks~\cite{qwen2025qwen25technicalreport}. In the experiment, the 7B model was selected for comparing GRPO and HVO, balancing performance and resource usage. In addition, we used GPT-4-turbo for GPT-4 and employed the already fine-tuned versions of PEGASUS on the BillSum\footnote{\href{https://huggingface.co/google/pegasus-billsum}{https://huggingface.co/google/pegasus-billsum}} and CNN/DailyMail\footnote{\href{https://huggingface.co/google/pegasus-cnn_dailymail}{https://huggingface.co/google/pegasus-cnn\_dailymail}} as baselines.

\subsubsection{Setup and Metrics}
For training the GRPO, the parameters are set as follows: train\_batch\_size = 64, num\_generations = 8, max\_grad\_norm = 0.4, and learning\_rate = 5e-7. All rewards are weighted equally with a weight of 1.0. Additionally, for HVO, we default to use $\epsilon$ = 0.99, $\delta$ = 0.1, $\rho$ = 16 and $\lambda$ = 2, $w^k$ is defaulted to -1 to align with GRPO. We primarily use UniEval (coherence, consistency, fluency, relevance, and the average of them) and hypervolume (HV) scores according to Equation~\ref{eq:hypervolume} \, as the evaluation metric.
\begin{figure*}
  \centering
  \includegraphics[width=\textwidth]{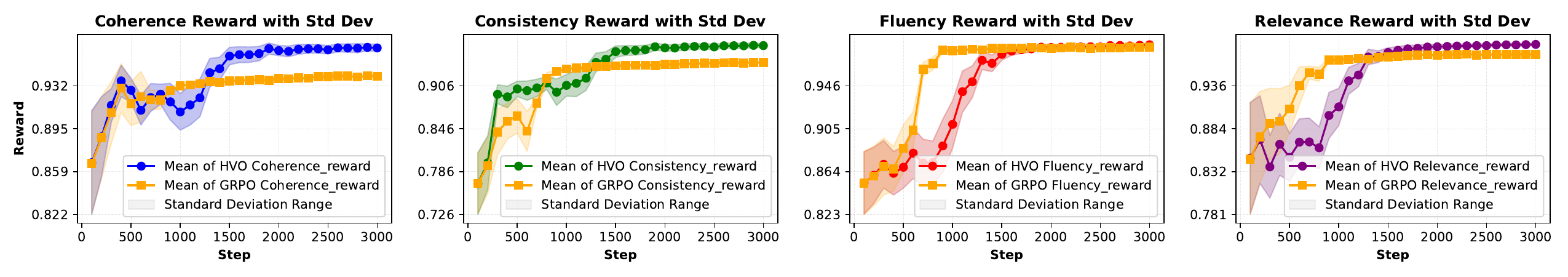}
  \vspace{-20pt}  
  \caption{The inter-group means and standard deviations on the 500-validation set during the training process on CNN/DailyMail, with HVO method recording scores prior to hypervolume calculation and being comparable to GRPO.}
  \label{fig:plot}
\end{figure*}
\subsection{Results}
\begin{figure}[ht]
  \centering
  \includegraphics[width=0.4\textwidth]{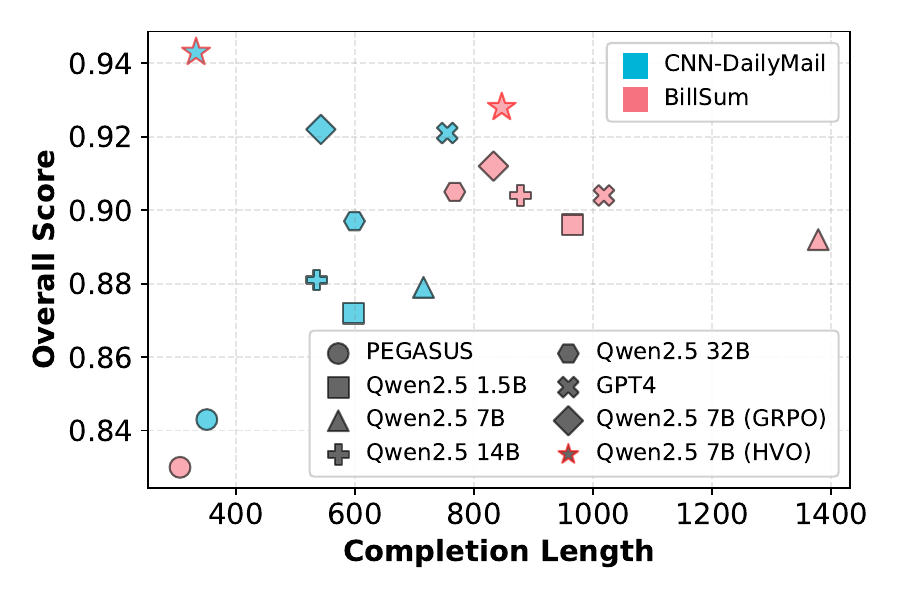}
  \caption{Scatter plot illustrating the relationship between the overall score and completion length, as evaluated by UniEval, for different models on two datasets: CNN/DailyMail and BillSum.}
  \label{fig:scattor}
  \vspace{-20pt}  
\end{figure}

Based on the experimental results from Table~\ref{tab:combined_results}, HVO outperforms all other methods in both datasets, demonstrating the highest hypervolume (HV) scores and overall scores. 

GPT-4, while excelling in coherence (0.967) and fluency (0.945) on the CNN/DailyMail dataset and coherence (0.973) and relevance (0.971) on the BillSum dataset, does not perform as well in overall performance and dimension balance compared to the Qwen 2.5 7B (HVO), which shows that the HVO method can achieve results comparable to GPT-4 in the summarization task. 

Though Qwen2.5 7B enhanced by both GRPO and HVO, shows balanced performance, HVO achieves much better overall scores. From Figure \ref{fig:plot}, in the early stages of training, the GRPO algorithm prioritizes fluency and relevance, with less emphasis on consistency. As a result, the optimization of consistency is constrained. In contrast, HVO optimizes all objectives more evenly and it is worth noting that the standard deviation of HVO is large and persists over a longer period of training, which makes the advantage signal more significant. This provides the opportunity to explore a larger strategy space, increasing the probability of approaching Pareto optimality, leading to more comprehensive and stable performance across all metrics.
The HV score serves as a proxy for how closely results approximate the pareto front. As shown in Table \ref{tab:combined_results}, HVO achieves a superior HV score. This indicates that our approach enables the policy model to approximate the pareto front more closely compared to existing baselines. 

Finally, we computed the generation length of different models and plotted a scatter plot of the overall score against completion length, as shown in Figure~\ref{fig:scattor}. Our method not only achieves the highest overall score but also maintains a shorter completion length, ensuring better conciseness.

\section{Conclusion}
In this paper, we introduce hypervolume optimization enhanced GRPO (HVO), a multi-objective reinforcement learning framework for text summarization that directly optimizes the hypervolume indicator in high-dimensional objective space. By balancing multiple evaluation metrics, HVO achieves a more stable and efficient trajectory toward the pareto frontier. Experiments on CNN/DailyMail and BillSum show that HVO attains state-of-the-art hypervolume and overall scores, outperforming existing methods and rivaling GPT-4, without supervised fine-tuning or cold-start initialization. These results confirm HVO’s effectiveness in managing complex trade-offs and generating high-quality summaries, offering a robust solution for multi-objective text summarization.




\label{ssec:subhead}
\label{sec:intro}
\bibliographystyle{IEEEbib}
\bibliography{strings,refs}

\end{document}